%% file: root.tex
\documentclass[conference]{IEEEtran}

\usepackage{amsmath,amssymb,amsfonts}
\usepackage{algorithmic}
\usepackage{graphicx}
\usepackage{textcomp}
\def\BibTeX{{\rm B\kern-.05em{\sc i\kern-.025em b}\kern-.08em
    T\kern-.1667em\lower.7ex\hbox{E}\kern-.125emX}}

\IEEEoverridecommandlockouts                              

\input{preamble}

\begin{document}

\title{The Finer Points:  A Systematic Comparison of Point-Cloud Extractors for Radar Odometry
\thanks{We acknowledge the support of the Natural Sciences and Engineering Research Council (NSERC) of Canada and the Ontario Research Fund-Research Excellence (ORF-RE) program. We would particularly like to thank Navtech for technical support of this work.}}
\author{\IEEEauthorblockN{Elliot Preston-Krebs, Daniil Lisus, and Timothy D. Barfoot}
\IEEEauthorblockA{\textit{University of Toronto Institute for Aerospace Studies (UTIAS)} \\
\textit{University of Toronto}\\
Toronto, Canada \\
Email: [FIRSTNAME].[LASTNAME]@robotics.utias.utoronto.ca}
}

\maketitle
\pagestyle{empty}

\sisetup{
propagate-math-font = true ,
reset-math-version = false
}

\begin{abstract}
A key element of many odometry pipelines using spinning frequency-modulated continuous-wave radar is the extraction of a point-cloud from the raw signal intensity returns. This extraction greatly impacts the overall performance of point-cloud-based odometry, but a consensus on which extractor performs best in which circumstances is missing. This paper provides a first-of-its-kind, comprehensive comparison of 13 common radar point-cloud extractors for the task of iterative closest point-based odometry in autonomous driving environments. Each extractor's parameters are tuned and tested on two FMCW radar datasets using approximately 176{\boldmath \, \unit{\km}} of data from public roads. We find that the simplest, and fastest extractor, \boldmath{$K$}-strongest, performs the best overall, outperforming the average by 13.59\% and 24.94\% on each dataset, respectively. In addition to an overall extractor recommendation, we highlight trends and note the substantial impact that the choice of extractor can have on the accuracy of odometry.

\end{abstract}


\section{INTRODUCTION}

In the context of autonomous vehicle navigation, spinning frequency-modulated continuous-wave (FMCW) radar offers robustness to adverse environmental conditions, a 360\textdegree \ view of its surroundings, long ranges, and penetrative properties, allowing for multiple returns along an azimuth \cite{mmWave,radarWeather}. Additionally, radar can perform well regardless of changes to lighting or time of day. Despite these advantages, spinning radar has limitations such as comparatively poor resolution, a detection plane limited to 2D, and susceptibility to structural and random noise in signals. Such impurities and artefacts can be observed as speckling, saturation lines, and `ghost' objects due to multi-path reflections. Additionally, many traditional radar filtering methods were developed for use in aircraft and sea-bound vessels, where the radar typically scans relatively sparse environments. This contrasts with the crowded urban settings where an autonomous vehicle is likely to operate. The combination of these factors leads to radar-based navigation in autonomous vehicles performing worse than state-of-the-art (SOTA) lidar navigation \cite{burnett2023readyradarreplacelidar}.

\definecolor{darkgreen}{rgb}{0.0, 0.5, 0.0}
\definecolor{darkred}{rgb}{0.7, 0.11, 0.11}

\begin{figure}[t]
    \centering

    \begin{tikzpicture}
        \node (photo) at (0,0) {\includegraphics[width=0.48\textwidth, trim={0 0cm 0 0cm},clip]{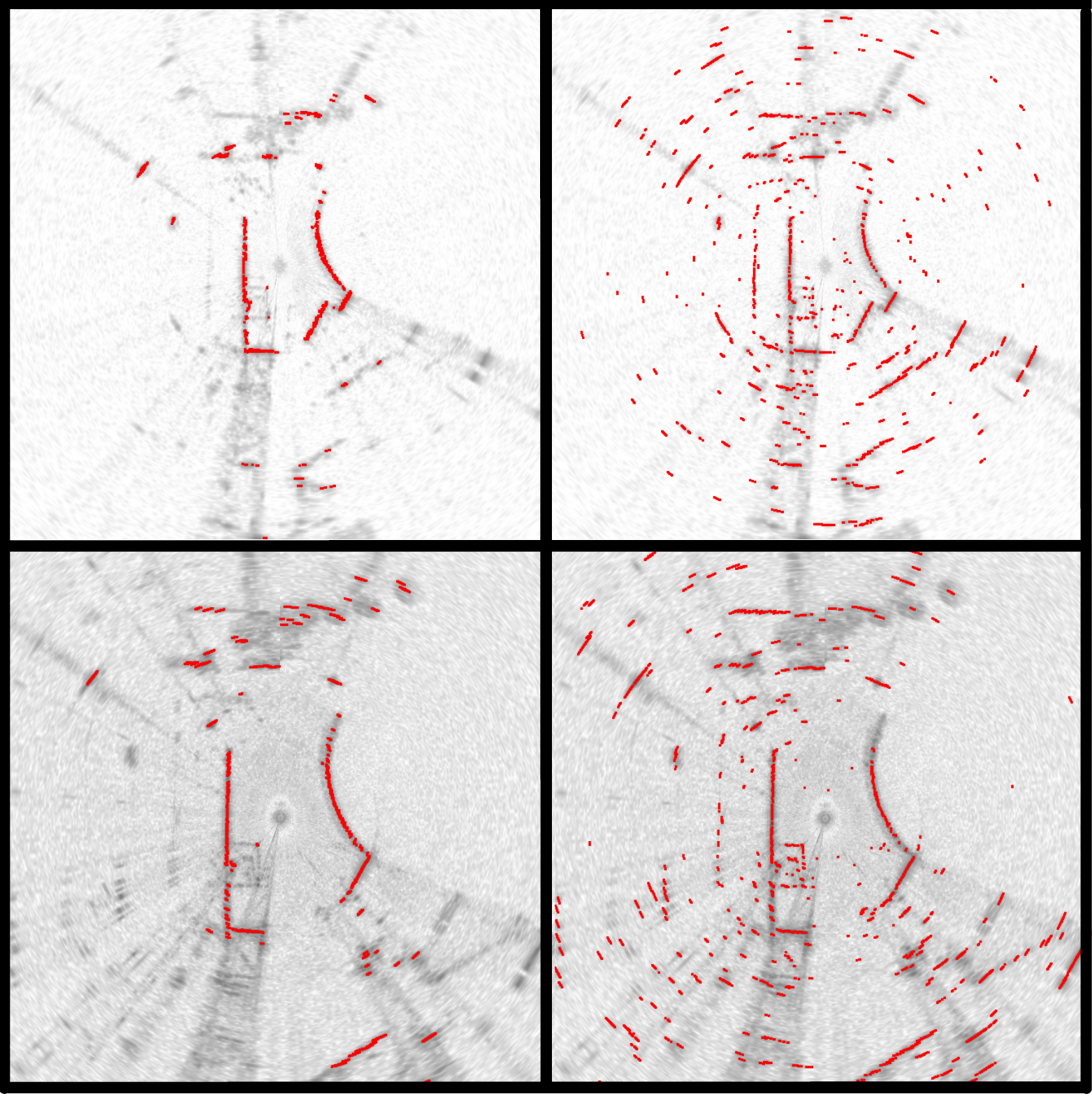}};

        \node[draw=black, rounded corners, fill=white, inner sep=0pt, line width=0.5mm] (photo) at (0,0) {\includegraphics[width=0.15\textwidth, trim={0 0cm 0 0cm},clip]{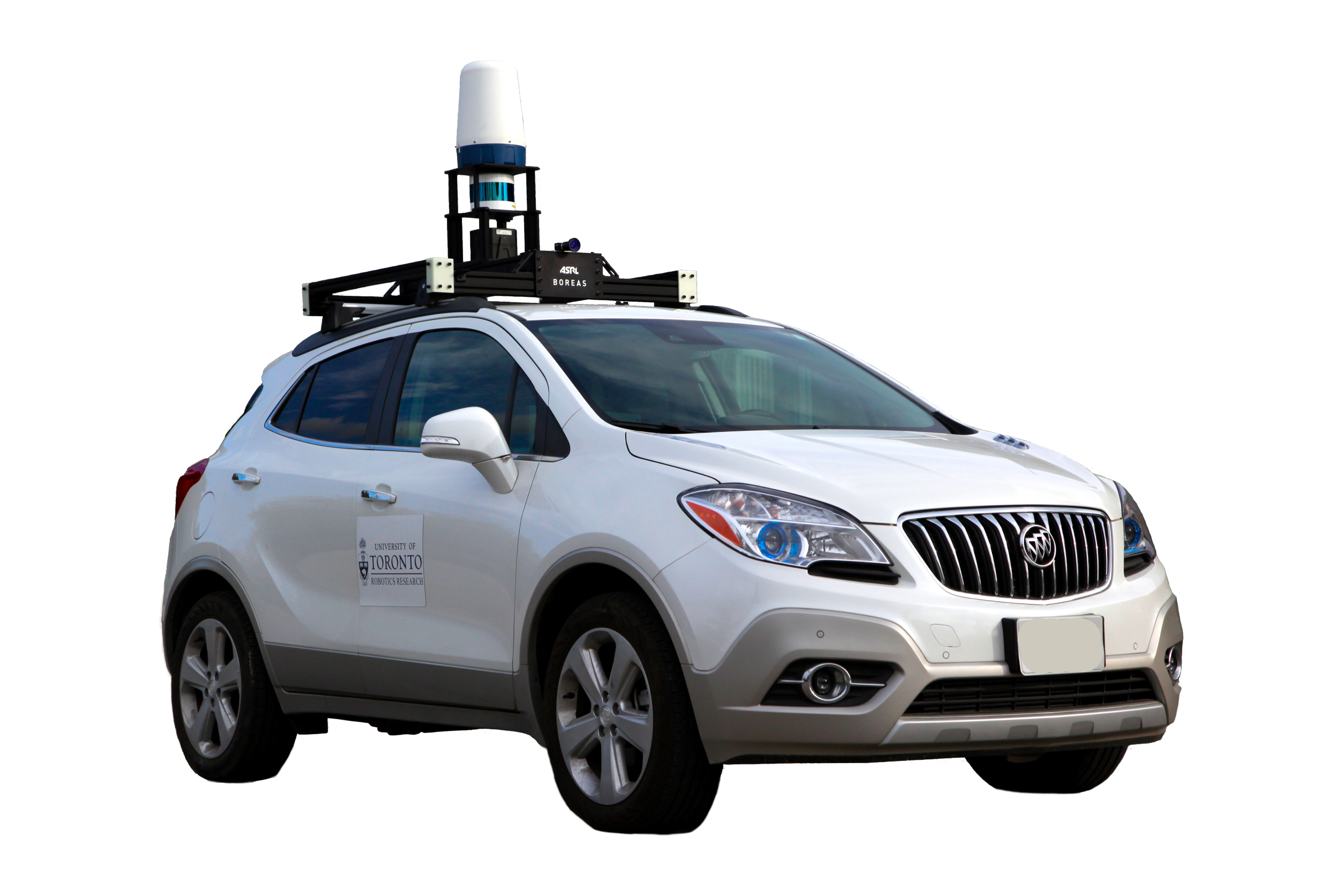}};

        \node[draw=darkgreen, rounded corners, fill=white, inner sep=5pt, line width=0.5mm] (best) at (-2.3,0.0) {{\color{darkgreen} \textbf{Best}}};
        \draw[-{Triangle[length=2.4mm,width=4.0mm]}, darkgreen, line width=1mm] (best) -- (-2.3,0.55);
        \draw[-{Triangle[length=2.4mm,width=4.0mm]}, darkgreen, line width=1mm] (best) -- (-2.3,-0.55);

        \node[draw=darkred, rounded corners, fill=white, inner sep=5pt, line width=0.5mm] (worst) at (2.3,0.0) {{\color{darkred} \textbf{Worst}}};
        \draw[-{Triangle[length=2.4mm,width=4.0mm]}, darkred, line width=1mm] (worst) -- (2.3,0.55);
        \draw[-{Triangle[length=2.4mm,width=4.0mm]}, darkred, line width=1mm] (worst) -- (2.3,-0.55);

        \node[draw=black, rounded corners, fill=white, inner sep=5pt, line width=0.5mm] (f1) at (0.0,3.92) {{\color{black} \textbf{F1 Dataset}}};
        \draw[-{Triangle[length=2.4mm,width=4.0mm]}, black, line width=1mm] (f1) -- (-1.25,3.92);
        \draw[-{Triangle[length=2.4mm,width=4.0mm]}, black, line width=1mm] (f1) -- (1.25,3.92);

        \node[draw=black, rounded corners, fill=white, inner sep=5pt, line width=0.5mm] (f2) at (0.0,-3.92) {{\color{black} \textbf{F2 Dataset}}};
        \draw[-{Triangle[length=2.4mm,width=4.0mm]}, black, line width=1mm] (f2) -- (-1.25,-3.92);
        \draw[-{Triangle[length=2.4mm,width=4.0mm]}, black, line width=1mm] (f2) -- (1.25,-3.92);

        \node at (-3.5,3.92) {\small \textbf{CFEAR}};
        \node at (-3.6,3.62) {\small {\color{darkgreen} \textbf{1.31\%}}};

        \node at (-3.28,-3.92) {\small \textbf{K-strongest}};
        \node at (-3.6,-3.62) {\small {\color{darkgreen} \textbf{1.03\%}}};

        \node at (3.05,3.92) {\small \textbf{CAGO-CFAR}};
        \node at (3.6,3.62) {\small {\color{darkred} \textbf{1.89\%}}};

        \node at (3.05,-3.92) {\small \textbf{MSCA-CFAR}};
        \node at (3.6,-3.62) {\small {\color{darkred} \textbf{1.49\%}}};

    \end{tikzpicture}
    \caption{Visualized point-clouds of the best- and worst-performing extractors and their percentage average translation errors. The top and bottom rows illustrate extractor performance in the same geometric environment under the differing noise characteristics of the two datasets collected using the Boreas platform, which is depicted in the centre. Gray indicates raw radar data, while red represents the extracted points.}
    \label{fig:A}
\end{figure}
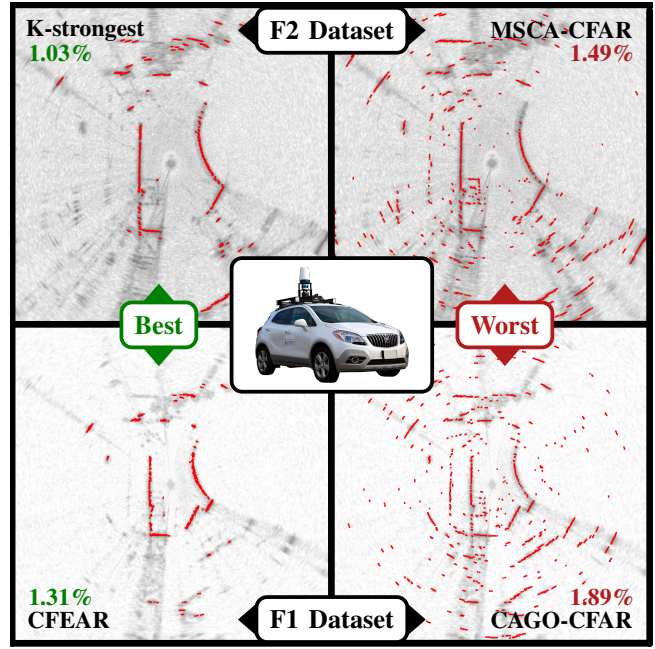
For scanning radar to be a more viable and competitive sensor for autonomous vehicle navigation, the filtering and processing of radar data must improve and become more consistent. Point-cloud-based algorithms generally represent the SOTA in spinning radar-based navigation \cite{burnett2023readyradarreplacelidar,lisus2024dopplervelocitymeasurementsuseful,burnett2021needcompensatemotiondistortion,lisus2024dopplervelocitymeasurementsuseful,lisus2024pointingwayrefiningradarlidar,CFEAR_OG,Adolfsson_2023,BFAR,CFEAR_Summer,cen2018,cen2019,ICP_indoors,ICP_Vehicles}. These algorithms all rely on the point-cloud extraction process, where a raw radar scan is refined into a set of points in Cartesian space. Despite the extractor being a critical component of radar-based systems, there has yet to be a comprehensive study comparing its impact on the performance of radar navigation in general. As a first step in evaluating the impact of extraction on radar navigation, this paper assesses 13 common extractors in the context of on-road radar odometry. The odometry pipeline we use to compare such extractors incorporates an iterative closest point (ICP) algorithm to match the point-clouds between scans \cite{burnett2023readyradarreplacelidar}. We have selected ICP as it relies solely on raw point-clouds generated by the various extractors, without requiring additional information such as descriptors. This direct dependence on point-clouds makes ICP more sensitive to variations between extractors, making it the most suitable odometry method for our comparison.

For this paper, we categorize the extractors into two groups: signal extractors and spatial extractors. Signal extractors process the radar scan azimuth-wise, while spatial extractors additionally use the neighbouring azimuths or a Cartesian representation of the radar scan to determine valid points. There exists an abundance of different extractors but a lack of comparison between them. `Which techniques work best and in which situations?' is still a question that needs to be answered in order to streamline the implementation and development of radar-based navigation.

The primary contribution of this paper is, to the best of our knowledge, the first thorough comparison of a curated list of several extractors for point-cloud-based radar-odometry. Our comparison sheds light on the magnitude of impact that extractor optimization has on overall performance, as well as the difference in extraction quality due to varying radar signal properties. Empirically, we observe that optimizing the radar front-end can yield greater improvements than optimizing the back-end of a given estimator, justifying the need for such a study to be conducted. The contributions of this paper begin to address a gap in the current literature on radar-odometry with the recommendation of an extractor for best performance in on-road environments, as well as insights into the behaviour and robustness of extractors when processing radar data of varying quality.

The rest of this paper is structured as follows. Section II outlines related work. Section III describes the extractors we compare as well as our tuning method. Section IV presents our data and corresponding results. Lastly, Section V provides a conclusion and recommendation.

\section{RELATED WORK}

\subsection{Radar Odometry}

The goal of radar odometry is to estimate the relative transform between consecutive radar frames; the estimates are then used to determine the overall path travelled by the vehicle. This is typically done using point-clouds, sets of features and descriptors, or the entire radar scan.

Point-matching algorithms determine the change in orientation and translation between frames by matching point-clouds, usually using ICP or a variant thereof \cite{burnett2021needcompensatemotiondistortion,lisus2024dopplervelocitymeasurementsuseful,cen2018}. 

Feature-matching algorithms rely on the extraction of features and descriptors using heuristic methods such as SURF or SIFT \cite{RADARSLAM,SLAMwVis}, or learning-based methods to predict key-points and pose estimates \cite{undertheradar,burnett2021radarodometrycombiningprobabilistic}.

Scan-matching algorithms approach odometry by performing correlative scan matching between entire scans instead of directly extracting and comparing points or features. Some scan-matching methods apply a Fourier-Mellin transform to a Cartesian or log-polar representation of the radar scan \cite{PhaRaO,SLAMwFMT}. Others have used machine learning to mask regions of a scan before using correlative matching \cite{maskbymove}. 

In our odometry pipeline, we use a continuous-time ICP algorithm, as we have found it to perform at or near the SOTA for radar-based odometry \cite{burnett2023readyradarreplacelidar,lisus2024dopplervelocitymeasurementsuseful}. Moreover, we use point-to-point ICP specifically because it relies solely on the quality of the individually extracted points, as opposed to the quality of secondarily extracted information used in point-to-line or point-to-plane cost functions such as normals. The characteristics of our selected odometry pipeline are conducive to the comparison of these extractors and the analysis of other types of estimators is left for future work.

\subsection{Signal Extractors}

Signal-based extractors discern points by exerting a threshold on the individual azimuths of a radar signal. The most prevalent method is the Constant False Alarm Rate (CFAR) extractor, initially designed to achieve a fixed probability of false alarm, or a false positive detection \cite{Finn_Johnson_1968}. The many variants of CFAR, of which there are more than 25 \cite{machado2017evaluation}, process each azimuth of a radar scan with a sliding window. The power readings inside this window are used to calculate a threshold for detection.

Variants such as Cell Averaging (CA)-CFAR \cite{CFAR_clutter}, Cell Averaging Greatest Of (CAGO)-CFAR \cite{CFAR_nonhomo_background}, Cell Averaging Smallest Of (CASO)-CFAR \cite{CFAR_nonhomo_background}, and Bounded False Alarm Rate (BFAR) \cite{BFAR} isolate portions of the sliding window to combat specific noise characteristics such as clutter edges and interfering targets, however fail to handle all types of non-homogeneous noise at once. Some extractors such as AND-CFAR \cite{ANDORCFAR}, OR-CFAR \cite{ANDORCFAR}, Improved Switching (IS)-CFAR \cite{ISCFAR}, and Variable Index (VI)-CFAR \cite{VICFAR} aim to supplement the weakness of the aforementioned extractors by having different modes of operation, conditional on the noise characteristics of the reference window. Other extractors such as Order Statistic (OS)-CFAR \cite{OSCFAR}, Mean Level Detector (MLD) \cite{MLD}, Censored Mean Level Detector (CMLD) \cite{CMLD}, Trimmed Mean (TM)-CFAR \cite{TMCFARbase,youheTMCFAR}, and Minimum Selected Cell Averaging (MSCA)-CFAR \cite{MSCACFAR} take a statistical approach by either selecting the cell with $k$th quantile strongest return in the reference window, or by pruning the outliers of a reference window to better estimate the average noise in a signal. 

Although CFAR extractors are the most prevalent, other extractors can be classified as signal-based. $K$-strongest \cite{CFEAR_OG} takes the $K$ strongest power returns above a static threshold along an azimuth. The point-cloud extractor presented by Cen et al. \cite{cen2018}, which is referred to as `C18' in this paper, applies several filters to the raw signal along an azimuth to isolate peaks and remove multi-path reflections. 

\subsection{Spatial Extractors}

A subset of spatial extractors leverage vision-based algorithms, such as SURF \cite{RADARSLAM} or SIFT \cite{SLAMwVis}, to extract features and descriptors from Cartesian representations of radar data. Others aim to extract features and descriptors using neural networks instead \cite{undertheradar,burnett2021radarodometrycombiningprobabilistic}. These types of extractors require both the features and descriptors to perform odometry, and because of this dependency, their performance cannot be evaluated within a traditional ICP pipeline. Another spatial extractor, proposed by Cen et al. \cite{cen2019} and referred to as `C19' in this paper, applies image filtering techniques to a polar image of a radar scan, targeting areas with high intensity and low gradients. The CFEAR radar odometry method, proposed by Adolfsson et al. \cite{CFEAR_OG,BFAR}, extracts an initial point-cloud with the previously mentioned $K$-strongest or BFAR extractor, but further filters these points based on low density and irregularities in local geometry. The surface normals and distributions of point clusters are then calculated and used in point-to-line odometry.

\subsection{Extractor Comparisons}
Typically, in papers proposing or analyzing CFAR extractors, emphasis is placed on the probabilistic characteristics of the extractors and their behaviour in environments with homogeneous and non-homogeneous noise, rather than on their performance in the context of a task such as radar odometry \cite{machado2017evaluation,CFAR_COMP_ALT,CFARcompalt}. This includes evaluating the performance of the extractors under Swerling I and II target models to simulate the different statistical behaviours of a radar return \cite{CFAR_clutter,CFAR_nonhomo_background}, as well as different types of background conditions such as Gaussian noise and Rayleigh clutter \cite{CFAR_COMP_ALT}. Papers discussing CFAR variants in the context of radar odometry often focus solely on popular methods such as CA-CFAR or BFAR, leaving many variants unexplored \cite{burnett2023readyradarreplacelidar,cen2018,BFAR}. Some papers dismiss CFAR methods altogether, deeming them too inconsistent \cite{burnett2021needcompensatemotiondistortion,PhaRaO}. Other papers simultaneously change the extractor and the back-end odometry pipelines, preventing a direct comparison of the extractors themselves \cite{CFEAR_OG,burnett2021radarodometrycombiningprobabilistic,BFAR}. 

There are currently no studies contrasting many different extractors in the context of radar-based navigation using a standardized pipeline. This paper begins to address this gap by providing a detailed comparison of 13 extractors in the context of on-road radar odometry. These extractors were selected based on their prevalence and the variety of methodologies they employ for point-cloud extraction. We aim to extend this work to off-road scenarios and for other navigation tasks, such as localization, in future research.


\section{METHODOLOGY}
\label{sec:METHODOLOGY}

\subsection{Odometry}
Our ICP-based odometry pipeline is based on the teach and repeat framework \cite{burnett2023readyradarreplacelidar,vtr_og}. The first stage of this odometry pipeline is a pre-processing module that uses an extractor to produce a point-cloud from the raw radar scan. This point-cloud is then fed into a continuous-time ICP algorithm using a white-noise-on-acceleration motion prior. Point-to-point ICP is used to optimize the transformation between the current point-cloud and a sub-map formed from several previous point-clouds. The output of this pipeline is an estimate of the change in 2D translation, 2D velocity, and heading between radar frames. 

For brevity and to present the most representative extractor comparison, we focus on point-to-point ICP as it is most dependent on the raw point-clouds, maximizing the impact of the extractor used. The choice of extractor is significantly more impactful than the choice of motion prior and the decision between discrete- or continuous-time state representation, allowing us to extrapolate the conclusions in this paper to other point-cloud-dependent odometry pipelines.

\subsection{Units of Power} 
\label{subsec:power}
The azimuths of a radar scan contain power intensities across the range bins, which are used to detect points. The choice of units of power that an extractor uses is a fundamental decision in their implementation. There is no consensus on a standard unit of power among extractors, and many papers lack clarity on which unit is used. 

For CFAR extractors, we converted the power signal to Watts despite the raw FMCW radar data being in \unit{\dB}. This is because estimating the average clutter power, a key component of CFAR extractors, using the arithmetic mean of power in Watts is more appropriate than averaging power values in \unit{\dB} on a logarithmic scale.

For other extractors, such as C18 and C19, we keep the units of power in \unit{\dB} to maintain authenticity to the original implementations \cite{cen2018,cen2019}. 

For $K$-strongest and CFEAR, we arbitrarily use \unit{\dB} as they only require the order of range bins by signal strength.

\subsection{Signal Extractors - CFAR Variants}
We tune and compare 11 signal extractors, of which 9 are CFAR variants. Extractors often encounter challenges such as non-homogeneous background clutter, clutter edges, and multiple interfering targets. Non-homogeneous clutter refers to noise in radar signals, which is non-uniformly distributed. Clutter edges, or boundaries, are portions of the signal where there is an abrupt change in the average clutter power. An interfering or multi-target environment is one where there are sections of a radar signal with more than one peak in close proximity \cite{CFAR_clutter,CFAR_nonhomo_background,CFAR_COMP_ALT}. CFAR extractors aim to handle these specific situations.

\begin{figure*}[t]
    \centering
    \includegraphics[width=0.645\textwidth]{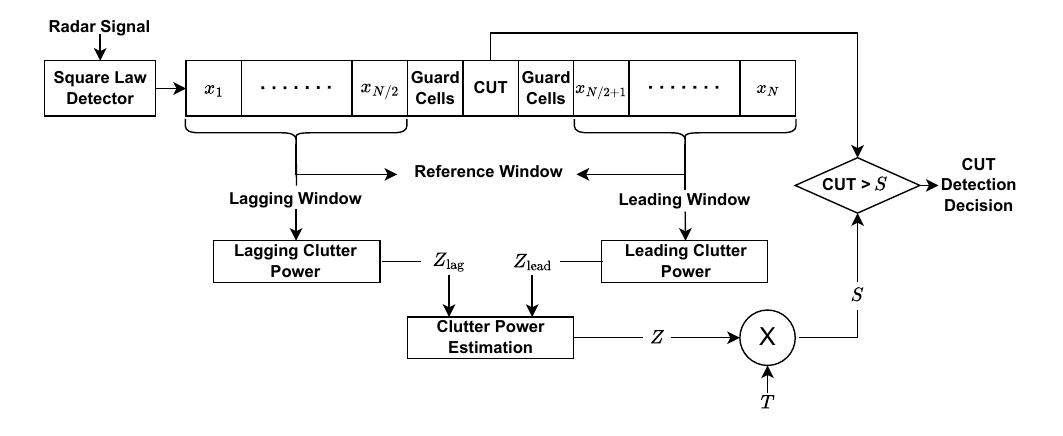}

    \caption{Generalized CFAR extractor schema. Diagram is zoomed in on a given reference window capturing cells $x_1$ to $x_N$.}
    \label{fig:CFAR}
\end{figure*}

CFAR variants operate along an azimuth, using the power readings within a sliding window, or reference window, to determine an appropriate threshold $S$. This threshold is then compared against the current center of the sliding window, or the cell-under-test (CUT), to determine if the CUT should be considered a point. CFAR extractors include guard cells, which are used to prevent the power of cells surrounding the CUT from influencing the average. For all CFAR variants, we square the input signal to simulate a square-law detector, a common component in almost all variants \cite{CFAR_clutter,CFAR_nonhomo_background,youheTMCFAR,ISCFAR,MSCACFAR,CMLD,VICFAR,OSCFAR,BFAR}. The generalized CFAR pipeline is visualized in Fig. \ref{fig:CFAR}.

The threshold $S$ is composed of an estimate of the average clutter power $Z$, and a scaling factor $T$ as 
\begin{align}
    S &= T\,Z. \label{eq:S}  
\end{align}
How the average clutter power is determined is unique to different extractors, while $T$ is derived from the probability of false alarm $P\textsubscript{fa}$. The basic definition of $T$ is computed as 
\begin{align}
    T &= N\,\left(P\textsubscript{fa}^{-1/N}-1\right),\label{eq:T} 
\end{align}
where $N$ is the reference window size \cite{CFAR_nonhomo_background}. The relationship between $P\textsubscript{fa}$ and $T$ can differ among extractors, and determining the optimal value for $P\textsubscript{fa}$ to maximize the probability of detection is a standard component of the theoretical analysis and tuning of CFAR extractors. However, for simplicity and consistency in our implementations, we directly tune $T$ instead of $P\textsubscript{fa}$, and fix the value of $N$ and the number of guard cells for all CFAR extractors.

\subsubsection*{1) CA-CFAR \cite{CFAR_clutter}} This extractor is the most basic of the CFAR variants, estimating $Z$ with an arithmetic mean of the power intensities in the reference window. Qualitatively, CA-CFAR experiences a decline in performance when there are interfering targets or in areas with sudden changes in background clutter power \cite{CFAR_nonhomo_background}.

\subsubsection*{2) CAGO-CFAR \cite{CFAR_nonhomo_background}} This extractor calculates the average clutter power by splitting the reference window into leading and lagging widows and using the larger of the two with $Z=\mathrm{max}(Z\textsubscript{lead},Z\textsubscript{lag})$. Selecting the window with the larger average improves detection when there is a clutter edge within the reference window. However, this choice conversely reduces probability of detection in multi-target scenarios \cite{CFAR_nonhomo_background,ISCFAR}.

\subsubsection*{3) CASO-CFAR \cite{CFAR_nonhomo_background}} This extractor selects the smaller average between the leading and lagging windows, where $Z=\mathrm{min}(Z\textsubscript{lead},Z\textsubscript{lag})$. CASO-CFAR handles closely spaced targets along an azimuth better compared to CA-CFAR and CAGO-CFAR, but fails to successfully detect points when there exists a clutter edge within the reference window.

\subsubsection*{4) IS-CFAR \cite{ISCFAR}} This extractor aims to reduce excessive false alarms caused by clutter edges and interfering targets by estimating clutter power through different switching cases. First, the intensities of the cells in each of the half-windows are compared to the CUT's intensity scaled by $\alpha$. If the cells are greater than the scaled CUT, they are considered an interfering target and added to the set $S\textsubscript{lead}$ or $S\textsubscript{lag}$. If the size of $S\textsubscript{lead}$ and $S\textsubscript{lag}$ both stay within a threshold $I$, then $Z$ is estimated using all cells not in $S\textsubscript{lead}$ and $S\textsubscript{lag}$. If the size of either $S\textsubscript{lead}$ or $S\textsubscript{lag}$ exceeds $I$, it indicates that the CUT is too close to the noise floor or a clutter edge. In this case, $Z$ is estimated using the half-window that violated $I$, including the interfering points to preserve the characteristics of those strong returns. If both sets exceed $I$, CA-CFAR is used including all interfering targets. This dynamic behaviour mitigates the impact that multi-target interference would have on CA-CFAR by pruning them from the clutter power estimation.

\subsubsection*{5) VI-CFAR \cite{VICFAR}} This extractor dynamically functions as CA-CFAR, CAGO-CFAR, or CASO-CFAR based on noise characteristics of the leading and lagging windows. The purpose of this is to provide robustness in homogeneous and non-homogeneous noise environments, which could include multiple targets and clutter edges. It uses two thresholds: the variability index threshold $V$ and the mean ratio threshold $R$. The variability index $V_i$ is calculated from each cell $x$ in either the leading or lagging window, as
\begin{align}
    V_i =  \frac{N}{2} \frac{\sum_{i=1}^{\frac{N}{2}} x_i^2}{\left(\sum_{i=1}^{\frac{N}{2}} x_i\right)^2} . \label{eq:VI}
\end{align}
If $V_i > V$, the noise environment in the respective window is classified as variable or non-homogeneous. If $V_i\leq V$, the noise environment is considered homogeneous. The mean ratio threshold $R$ is used to bound the similarity of the means of the leading and lagging window using
\begin{align}
    \frac{1}{R} < \frac{Z\textsubscript{lead}}{Z\textsubscript{lag}} < R, \label{eq:MR}
\end{align}
where the means are considered different if the ratio is outside of these bounds. If the two windows are both considered homogeneous and the means are similar, or if only one window is homogeneous, CA-CFAR is used on the homogeneous window(s). If the windows are both homogeneous, but have different means, this is considered a clutter edge and CAGO-CFAR is used. Lastly, if both windows have non-homogeneous noise, then CASO-CFAR is used. The original paper uses two different $T$ values: one for the case when only one half-window is used, and another for when both half-windows are used. For simplicity, our implementation employs a single scaling factor $T$ to address both scenarios. Additionally, we fix $R$ based on the values suggested in the original paper.

\subsubsection*{6) OS-CFAR \cite{OSCFAR}} This extractor takes a statistical approach to determining the average clutter power. Instead of calculating the mean clutter power in the reference window, OS-CFAR orders the cells in the window by their power readings, and selects the cell from the desired $k$th quantile of the list. In our implementation, we select the median power reading as the clutter power $Z$.

\subsubsection*{7) TM-CFAR \cite{TMCFARbase}} This extractor is a generalization of CA-CFAR and OS-CFAR. TM-CFAR orders the cells in the reference windows, trims the $N_1$ smallest values and $N_2$ largest values, and then calculates the average power of the trimmed window. In our implementation, we trim the entire sorted reference window as opposed the the leading and lagging windows separately so that fewer inliers are trimmed. The unique parameters are $N_1$ and $N_2$, which we set equal to simplify tuning and denote as $N_\textsubscript{T}$. By pushing $N_1$ and $N_2$ to the extremes, we can generalize TM-CFAR as CA-CFAR in the case no cells are trimmed, and OS-CFAR in the case that all cells except the median are trimmed.

\subsubsection*{8) MSCA-CFAR \cite{MSCACFAR}} This extractor uses a sub-reference window (SRW) with size $M$ inside the main reference window. This SRW compares the two cells at its edges, keeping only the minimum cell, and repeats this process through the entire reference window. All of these minimum cells are then used to calculate $Z$. By selecting the minimum value in this fashion, MSCA-CFAR aims to improve performance in non-homogeneous backgrounds with interfering targets.

\subsubsection*{9) BFAR \cite{BFAR}} This extractor calculated the average clutter power and has a scaling factor equivalent to CA-CFAR. The only difference is the addition of a static threshold $b$ to the overall threshold calculation $S = T\,Z + b$. The fixed-level threshold filters a baseline level of noise, and when combined with the scaling factor, creates a more sensitive and accurate extractor, compared to CA-CFAR, when properly tuned.

\subsection{Signal Extractors - Non-CFAR Variants}
\subsubsection*{10) K-Strongest \cite{CFEAR_OG}} This extractor selects the $K$ most intense power readings along an azimuth. Additionally, a static threshold $z\textsubscript{min}$ filters out weak returns, speeding up processing by reducing the number of cells to sort and removing noise when fewer than $K$ strong returns are present.

\subsubsection*{11) C18 \cite{cen2018}} This extractor transforms the radar signal by unbiasing with a median filter, smoothing it with a binomial filter, and then attenuating both the high- and low-frequency noise. The transformed signal is then compared to a threshold composed of the local noise and scaling factor $z\textsubscript{q}$. Additionally, there is a final optional step in the algorithm to remove multi-path reflections by comparing the wavelet transforms of the different peaks, and removing the secondary peaks that resemble the primary ones. C18 has three main parameters: the width of the median filter $w\textsubscript{median}$, the width of the binomial filter $w\textsubscript{binom}$, and the previously mentioned $z\textsubscript{q}$. Intuitively, $w\textsubscript{median}$ estimates the distance spanning multiple landmarks and $w\textsubscript{binom}$ represents the average width of a peak. There is a fourth parameter $d\textsubscript{thresh}$, which pertains to filtering multi-path reflections. In our implementation, we subtract the mean power from each azimuth instead of using a median filter in order to remove bias from the signal. Additionally, we replace the binomial filter with a Gaussian filter, and do not remove the multi-path reflections as it increases operation time and is considered optional. These simplifications significantly reduce processing time and the number of parameters that must be tuned. 


\newcolumntype{?}{!{\vrule width 1.2pt}}

\renewcommand{\arraystretch}{1.15}
\begin{table*}[ht]
    
    \centering   
    \scriptsize 
    \setlength{\arrayrulewidth}{0.25mm}
    \caption{Average translational \& rotational errors for each extractor, calculated across all specified sequences. The table lists the runtime and the average number of extracted points. The top rows pertain to the F1 and F2 datasets. Only tuned parameters are shown. Detectors marked with an asterisk are CFAR variants and extractor names are abbreviated.}
        \scalebox{0.93}{
            \begin{tabular}{?c|c|c?c?c|c|c|c|c|c?c|c?}
                \hline    
                
                \hline   
                \rule{0pt}{1.0ex}
                \multirow{2}{*}{} &\multirow{2}{*}{Extractor} & \multirow{ 2}{*}{Parameters} & \textcolor{blue}{ATE(\%)/ARE}& \multirow{ 2}{*}{\textcolor{blue}{2020-12-04}} & \multirow{ 2}{*}{\textcolor{blue}{2021-01-26}} & \multirow{ 2}{*}{\textcolor{blue}{2021-02-09}} & \multirow{ 2}{*}{\textcolor{blue}{2021-03-09}} & \multirow{ 2}{*}{\textcolor{blue}{2021-06-29}}& \multirow{ 2}{*}{\textcolor{blue}{2021-09-08}} & Avg Run-  & Avg \# \\
                & & & \textcolor{blue}{($10^{-3}$deg/m)} & & & & & & & time(ms) & Points  \\
                
                \hline
                
                \hline
                    
                \multirow{14}{*}{\begin{sideways}\textbf{F1 Dataset}\end{sideways}}& CA* \cite{CFAR_clutter} & T=35 &\textcolor{blue}{1.87/5.57}& 2.26/6.25&2.24/6.78&1.46/4.63&2.05/6.12&1.62/4.84&1.57/4.82&25.04&863 \\
                \cline{2-12}        
                 & CAGO* \cite{CFAR_nonhomo_background} & T=25     &\textcolor{blue}{1.89/5.62}&2.53/6.94&2.06/6.36&1.63/5.24&2.01/5.88&1.55/4.74&1.57/4.54&25.21&919\\
                \cline{2-12}        
                 & CASO* \cite{CFAR_nonhomo_background} & T=400& \textcolor{blue}{\textbf{1.29}/3.96}&1.47/4.26&1.40/4.53&1.24/4.15&1.13/3.41&1.16/3.30&1.32/4.13& 25.42&1203\\
                \cline{2-12} 
                 & IS* \cite{ISCFAR} &T=15, $\alpha$=0.075
                &\textcolor{blue}{1.84/5.59}&2.46/6.84&2.28/7.27&1.76/5.59&1.61/4.88&1.48/4.46&1.45/4.52&63.95& 1046\\
                \cline{2-12} 
                 & VI* \cite{VICFAR} &T=400, $V$=5&\textcolor{blue}{1.32/4.08} &1.68/4.75&1.38/4.53&\textbf{1.15}/3.98&1.27/3.74&1.18/3.40&1.26/4.09& 47.17& 1208\\
                \cline{2-12} 
                 & OS* \cite{OSCFAR} & T=120     &\textcolor{blue}{1.47/4.46} &2.01/5.43&1.68/5.20&1.30/4.36&1.26/3.83&1.20/3.73&1.39/4.20&124.24&1265 \\
                \cline{2-12} 
                 & TM* \cite{TMCFARbase}& T=100, $N\textsubscript{T}$=30 &\textcolor{blue}{1.62/4.85} &2.42/6.61&1.96/6.04&1.41/4.59&1.28/3.88&1.31/3.88&1.33/4.11&164.01&1020\\
                \cline{2-12} 
                & MSCA* \cite{MSCACFAR} & T=100, M=8 &\textcolor{blue}{1.53/4.85} &2.12/5.71&1.66/5.31&1.24/4.11&1.40/4.33&1.32/4.02&1.46/4.53& 40.18& 1091 \\
                \cline{2-12} 
                &BFAR \cite{BFAR}&T=15, b=19.13\protect\footnotemark[1]  &\textcolor{blue}{1.47/4.40} &1.80/4.95&1.82/5.63&1.56/4.92&1.39/4.17&1.11/3.23&1.13/3.49& 24.27&868\\
                \cline{2-12} 
                 & K-str \cite{CFEAR_OG} & k=5 &\textcolor{blue}{1.37/4.07} &1.68/4.86&1.87/5.38&1.38/4.36&1.30/4.04&\textbf{0.95/2.80}&\textbf{1.03/2.97}& \textbf{20.71}& 1498\\ 
                \cline{2-12} 
                 & C18 \cite{cen2018}&$w\textsubscript{b}$=10,$z\textsubscript{q}$=2.75  &\textcolor{blue}{1.81/4.26}&1.83/4.04&1.68/3.73&1.90/4.23&1.30/2.68&2.48/6.81&1.65/4.07& 38.36& 604\\
                \cline{2-12} 
                 & C19  \cite{cen2019} & $l\textsubscript{max}$=400 &\textcolor{blue}{1.82/5.08} &1.43/4.03&1.66/4.58&1.25/3.72&1.22/3.34&2.80/7.69&2.55/7.12& 89.45& 350\\
                \cline{2-12} 
                 & CFEAR \cite{CFEAR_OG}& k=20, r=0.5 &\textcolor{blue}{1.31/\textbf{3.81}}&\textbf{1.02/3.02}&\textbf{1.21/3.20}&1.25/\textbf{3.62}&\textbf{1.07/3.07}&1.59/4.76&1.74/5.22& 50.29& 445\\
                \cline{2-12}
                & \multicolumn{2}{c?}{Average Performance} &\textcolor{RoyalPurple}{1.59/4.66}&\textcolor{RoyalPurple}{1.90/5.21}&\textcolor{RoyalPurple}{1.76/5.27}&\textcolor{RoyalPurple}{1.43/4.42}&\textcolor{RoyalPurple}{1.41/4.11}&\textcolor{RoyalPurple}{1.52/4.44}&\textcolor{RoyalPurple}{1.50/4.45}&\textcolor{RoyalPurple}{56.79}&\textcolor{RoyalPurple}{952}\\
                \hline
                
                \hline     
                
                \multicolumn{3}{?c}{} &  & \textcolor{blue}{2021-11-02} & \textcolor{blue}{2021-11-06} & \textcolor{blue}{2021-11-14} & \textcolor{blue}{2021-11-16} & \textcolor{blue}{2021-11-23}& \textcolor{blue}{2021-11-28} & \multicolumn{2}{c||}{} \\
                \hline
                
                \hline
                
                \multirow{14}{*}{\begin{sideways}\textbf{F2 Dataset}\end{sideways}} &CA* \cite{CFAR_clutter} & T=55 &\textcolor{blue}{1.33/4.01} & 1.11/3.45&1.55/4.30&1.70/5.05&1.17/3.64&1.11/3.44&1.34/4.20& 35.81 & 1023 \\
                \cline{2-12}
                &CAGO* \cite{CFAR_nonhomo_background} & T=50 & \textcolor{blue}{1.34/3.98} & 1.17/3.44 &1.54/4.25&1.49/4.46&1.36/4.18&0.98/2.99&1.51/4.54&36.98& 926\\
                \cline{2-12}
                &CASO* \cite{CFAR_nonhomo_background} & T=3700 &\textcolor{blue}{1.39/3.91} & 1.22/3.37&1.58/4.27&1.51/4.23&1.38/3.82&1.29/3.84&1.38/3.95& 39.86&859\\
                \cline{2-12}
                &IS* \cite{ISCFAR} & T=5, $\alpha$=0.003 &\textcolor{blue}{1.43/4.28}& 1.29/3.94&1.42/4.28&1.45/4.18&1.51/4.43&1.23/3.84&1.66/4.99&94.06&1061\\
                \cline{2-12}
                &VI* \cite{VICFAR} & T=2000, $V$=5 &\textcolor{blue}{1.48/4.25}& 1.14/3.40&1.63/4.66&1.52/4.24&1.55/4.28&1.49/4.33&1.54/4.57&61.28&1129\\
                \cline{2-12}
                &OS* \cite{OSCFAR} & T=1000    &\textcolor{blue}{1.31/3.91}& \textbf{1.00/3.03}&1.59/4.68&1.37/4.027&1.35/4.03&1.18/3.58 &1.35/4.14&181.13&989\\
                \cline{2-12} 
                &TM* \cite{TMCFARbase}& T=1050, $N\textsubscript{T}$=44 &\textcolor{blue}{1.31/3.93}& 1.27/3.74&1.66/4.81&1.24/3.63&1.11/3.42&1.16/3.70&1.41/4.27& 221.88&953\\
                \cline{2-12}
                &MSCA* \cite{MSCACFAR} & T=400, M=10 &\textcolor{blue}{1.49/4.58}& 1.35/4.29&1.52/4.50&1.46/4.40&1.61/4.89&1.67/5.05&1.35/4.35&54.37&1133 \\
                \cline{2-12}
                &BFAR \cite{BFAR}& T=12.5, b=38.25\protect\footnotemark[1]  &\textcolor{blue}{1.24/3.67} & 1.12/3.36&1.08/3.07&1.45/4.29 &1.19/3.47&1.24/3.71&1.36/4.11 &35.29&1033 \\
                \cline{2-12}
                &K-str \cite{CFEAR_OG} & k=3 & \textcolor{blue}{\textbf{1.03/2.85}} & 1.15/3.25&\textbf{1.03/2.77}&\textbf{1.04/2.79}& \textbf{0.93}/2.67 &1.00/2.71 &1.05/2.90& \textbf{26.27}& 750\\ 
                \cline{2-12}
                &C18 \cite{cen2018}&$w\textsubscript{b}$=6,$z\textsubscript{q}$=2 &\textcolor{blue}{1.11/3.02} & 1.18/3.29& 1.23/3.34&1.34/3.50&\textbf{0.93/2.59}&\textbf{0.97/2.62}&\textbf{1.04/2.78}&53.43&803\\
                \cline{2-12}
                &C19  \cite{cen2019} & $l\textsubscript{max}$=300 &\textcolor{blue}{1.87/5.37} & 2.24/6.62 &1.79/5.18 &1.80/5.07&1.91/5.39&1.58/4.51 &1.90/5.43 &263.37&251\\
                \cline{2-12}
                &CFEAR \cite{CFEAR_OG}& k=20, r=0.5 &\textcolor{blue}{1.51/4.36} & 1.40/4.27&1.61/4.55&1.39/3.88&1.62/4.71&1.39/3.91&1.64/4.81&64.44&439\\
                \cline{2-12}
                & \multicolumn{2}{c?}{Average Performance} &\textcolor{RoyalPurple}{1.37/4.01}&\textcolor{RoyalPurple}{1.28/3.80}&\textcolor{RoyalPurple}{1.48/4.20}&\textcolor{RoyalPurple}{1.44/4.13}&\textcolor{RoyalPurple}{1.36/3.96}&\textcolor{RoyalPurple}{1.25/3.71}&\textcolor{RoyalPurple}{1.43/4.23}&\textcolor{RoyalPurple}{89.86}&\textcolor{RoyalPurple}{873}\\
                \hline
                
                \hline
            \end{tabular}
        }
    \label{tab:old_results}
\end{table*}

\subsection{Spatial Extractors}
\subsubsection*{12) C19 \cite{cen2019}} This extractor processes an image of the polar radar scan by applying the Prewitt operator to calculate its intensity gradients. These gradients are then multiplied with the input image to accentuate regions with high intensity and low gradients, which typically correspond to consistent features. The intensities resulting from this multiplication are sorted in descending order. For each selected intensity, a region is defined along the respective azimuth, encompassing neighbouring cells with high intensity. The number of masked regions is controlled by the parameter $l\textsubscript{max}$. The extractor iterates through these regions, selecting the highest intensity point in each region only if there are neighbouring regions on adjacent azimuths. This ensures that the chosen point represents a consistent feature. This final set of points is then returned as a point-cloud in Cartesian space.
\subsubsection*{13) CFEAR \cite{CFEAR_OG}} This extractor was originally created for a point-to-line or point-to-distribution odometry pipeline. The first step is to run $K$-strongest and collect a point-cloud with the points converted to Cartesian coordinates. Next, a grid with side lengths of $r/f$ is applied on a Cartesian representation of the scan, where $r$ is the expected radius in which enough points can be found to estimate a point normal, and $f$ is a re-sampling factor to modify the density of the final feature set. For each grid square center, the neighbouring points within a radius of $r$ are collected and further filtered based on if there are too few points, or if the points are lined up along an azimuth. The original CFEAR method computes the normals, means, and covariance of these point clusters to be used in point-to-line calculations. For our implementation, we use the mean location of each cluster, which is added to the final point-cloud.

\section{EXPERIMENTAL RESULTS}

\subsection{Dataset}
The Boreas dataset \cite{BOREAS} is used for training and testing the extractors. We use 22 sequences collected with the Navtech RAS6 radar sensor, spinning at 4\unit{\Hz}, and providing 400 azimuth measurements per rotation. The ground truth for these sequences is obtained from GNSS, IMU, and wheel encoder data, post-processed using Applanix's RTX POSPac software. The raw radar data provided by the sensor is a normalized 8-bit half-\unit{\dB} intensity in polar form. To prepare this data for the extractors, we convert the normalized half-\unit{\dB} signal into \unit{\dB}. Then as outlined in Section \ref{subsec:power}, we convert the signal to Watts using $P_{\unit{\watt} }=10^{P_{\unit{\dB}}/10}$, or leave it in \unit{\dB} depending on the extractor. Across these sequences, two distinct radar firmware versions were employed. Out of the 22 sequences, the first 12 were captured with a range resolution of $0.0596\,\unit{\m}$ per range bin, while the remaining 10 sequences used a resolution of $0.0438\,\unit{\m}$. In addition to the finer resolution, the second firmware also had a higher noise floor as can be seen in Fig. \ref{fig:A}. Since the extractors are highly dependent on radar-specific parameters, including the range resolution and noise characteristics, we independently tune them for each dataset to account for the variation in radar attributes. As a result, we treat the sequences from the respective firmware versions as two distinct datasets denoted F1 and F2. This independent tuning reveals more general trends in performance, exposing extractors that were more or less robust to varying radar characteristics.

\footnotetext[1]{Values are displayed in \unit{\dB}. Squared Watts are used in calculation.}

\subsection{Fixed Parameters}
For all applicable CFAR extractors, we used 5 guard cells on each side of the CUT and a reference window size of 100 cells. These values were selected based on preliminary tuning results to standardize the extractors. Some of the selected extractors had more than two adjustable parameters, so to avoid excessive, multi-dimensional tuning, we performed preliminary experiments guided by recommendations from the literature to determine which parameters did not significantly impact the performance of an extractor and could remain fixed. The values of the fixed parameters were selected based on repeatedly produced basins of performance across several sweeps of an extractor's other parameters. By fixing parameters with a smaller impact on performance, we were able to conduct more rigours fine-tuning of these extractors, resulting in more definitive basin's of optimality.

For TM-CFAR, we used only one parameter to trim the sorted reference window $N_\textsubscript{T}$, which would trim the same number of cells from the top and bottom quantiles of the sorted window. For VI-CFAR we fixed $R=1.5$ based on recommended values \cite{VICFAR}. For IS-CFAR we did preliminary tuning on the parameter $I$, which pertains to the tolerable number of interfering targets in the reference window and found $I=6$ worked the best. For $K$-strongest and CFEAR, since it uses $K$-strongest, we used a fixed threshold of $z_\textsubscript{min}=31.875\, \unit{\dB}$ in F1 and $z_\textsubscript{min}=44.625 \,\unit{\dB}$ in F2, which we selected based on preliminary testing and engineering judgment. For CFEAR, we tuned the re-sampling factor $f$ indirectly by experimenting with various grid sizes based on $r/f$. We determined that a grid size of $r/f=0.5$ was effective from visual inspection and preliminary metrics. For tuning, we fixed this grid size and adjusted $r$ separately. 

\subsection{Tuning \& Evaluation}
When tuning the extractors, we followed a general protocol: first identifying a basin of optimal performance with a coarse parameter sweep, then performing a finer sweep to find the optimal parameters. We measured performance using the KITTI odometry metrics with the percent average translational error (ATE) and average rotational error (ARE) in (\unit{\deg\per\m}) averaged over $100$, $200$, \ldots, $800\,\unit{\m}$ path lengths \cite{KITTI}. Optimal parameters were selected solely based on the minimum ATE across the training sequences. These parameter configurations were then used on the testing sequences to produce a final ATE and ARE.

The approach of tuning the extractors with the ATE was adopted due to the high correlation in performance between the two metrics. During coarse sweeps, we observed that the basins of convergence for ATE and ARE were highly correlated, with no instances where a set of parameters would result in exceptional ARE but poor ATE, or vice versa. Given that the ARE was consistently small across all extractors, and thus already generally acceptable for odometry, we focused on optimizing the ATE, which is the more critical metric for improving overall performance. This decision was further supported by the fact that the ATE is inherently impacted by the rotational accuracy, since the orientation affects the estimated displacement between two positions.

As an additional comparison metric, we measured both the single-thread per-frame average extractor runtime and the average number of points extracted by each extractor. The average number of extracted points provides an algorithm-agnostic heuristic metric for the relative runtime expected from a point-cloud-based pipeline. Depending on the algorithm and available computational resources, a highly accurate algorithm may extract a greater number of points and thus fail to run in real time. However, since the notion of `real-time performance' varies between platforms, applications, and algorithms, we only consider the final accuracy when selecting the `best' extractors in this paper.

Out of the 12 sequences in F1, 6 were used for tuning and 6 for testing. For the 10 sequences in F2, 4 were allocated for tuning and 6 for testing.

\subsection{Results}
Table \ref{tab:old_results} presents the results from all considered extractors across both datasets. 

\subsubsection{Signal Extractors}
Most signal extractors showed a high degree of variance between the datasets. Extractors such as VI and C18 excelled in F1 and F2, respectively, but showed poor performance when tested on the other dataset. Within F1, some CFAR variants, such as CASO and VI, were among the top performers, while others, such as CA and CAGO, were at the bottom; in F2, all CFAR variants performed within a similar range when optimized.

We hypothesize that this inconsistency results primarily from the different firmware characteristics, with F1 having a lower noise floor but more diffuse geometry than F2. Diffuse geometry, as a result of a coarser range resolution, contains more interfering targets. Consequently, extractors that directly target this type of challenge, for example CASO and VI-CFAR, stand out in performance much more when tested on F1 than F2. However, a finer range resolution generally benefited all forms of signal extractors, as the increased number of cells for the same range permits more accurate signal characteristic computations. This is particularly highlighted for the C18 extractor, which is one of the worst performers in F1 and is one of the best in F2. C18 performs multiple rounds of smoothing and filtering, most of which are based on computed signal characteristics that would be expected to improve with a higher range resolution. Interestingly, the variance of CFAR-based extractors decreased significantly in F2. Most variants are designed to better handle specific types of artefacts and noise, many of which (e.g. interfering targets) become less challenging as the range resolution improves. 

Overall, a finer range resolution tends to provide better performance for signal extractors, as indicated by the lower average ATE and the smaller average number of points required to achieve that performance in F2. However, this improvement comes at the cost of longer extraction times, as the finer resolution increases the number of range bins to process. This must be considered when balancing accuracy with computational efficiency, especially in real-time applications.

\subsubsection{Spatial Extractors}
In contrast to signal extractors, diffuse geometry benefited spatial extractors, with an observed performance drop from F1 to F2. We speculate that more diffusion in the signal provides better spatial information and understanding, improving the robustness of spatial methods that rely on more extensive scene information. Although spatial extractors performed on a similar average level to signal extractors in F1, with CFEAR even being a top extractor, they got significantly outperformed by signal extractors in F2, likely owing to the benefits that signal extractors derived from a finer range resolution.

However, it should be noted that spatial extractors tend to be more algorithm-specific as compared to signal extractors. For example, we note that our point-to-point-based CFEAR implementation does not reach the current SOTA radar odometry performance that the original CFEAR radar odometry pipeline does \cite{CFEAR_Summer}. This is likely due to CFEAR being designed to extract sparser information, such as normals and surface points, to be used with point-to-line or point-to-distribution ICP specifically. Conversely, although C19 does quite poorly on both datasets when compared to other extractors, the original algorithm specifically aims to perform well in challenging off-road environments. As such, C19 may be a much better candidate when faced with environments for which it was designed for. This highlights that tailored extractors can certainly outperform more `general' ones when design decisions are made with a specific algorithm or environment in mind.

\subsubsection{Overall Recommendations}
Across both datasets, $K$-strongest consistently ranked among the top performers, achieving an ATE of 1.37\% on F1 and 1.03\% on F2. These results were 13.59\% and 24.94\% better than the average ATE for the two datasets, with $K$-strongest securing a dominant first-place performance on F2. $K$-strongest also had the fastest runtime across both datasets due to its simple design. However, it extracted the most points on average in F1, which may result in a longer overall algorithm runtime. Despite this, the high degree of control over the number of points extracted with the single variable $K$ could be used to compensate for this. As seen in F2, $K$-strongest was below the average of the number of points extracted due to a lower $K$. If algorithm runtime was a bottleneck on a real-time system, $K$ could be decreased to a lower value, typically without a significant loss of performance. Therefore, we recommend the $K$-strongest approach as a generally reliable extractor due to its high accuracy relative to other extractors, its simplicity, and its ease of tuning for both performance and runtime. However, this does not mean that $K$-strongest will always outperform other extractors on a given dataset, in a given environment, or when coupled with a given estimation algorithm. If the absolute best performance is desired, hand-tuning or even hand-crafting an extractor will likely yield improvements.

Finally, this comparison highlights the impact that the extraction process can have on the overall estimation result. The average estimation quality in our comparison could be improved by 46.51\% in F1 and 81.55\% in F2 simply by swapping out the worst-performing extractor for the best. Such an improvement is many times more impactful than most improvements in back-end processing approaches. Therefore, it is critical to consider front-end improvements in tandem with back-end development in order to make radar a more competitive sensor to lidar and cameras in all weather conditions.

\subsection{Limitations \& Extensions}
This paper provides a basis of comparison for the aforementioned extractors, specifically in the context of radar-based odometry on public roads. While these conclusions can inform the selection of a point-cloud extractor for mobile robots and self-driving vehicles, there is still room for further experimentation and comparison of radar point-cloud extractors used by robots navigating in other environments. Field robots traversing off-road terrain or mobile robots navigating the inside of a building would be faced with different environmental factors influencing the quality or characteristics of the raw radar signal, and as a result, could be better suited with another point-cloud extractor. In addition to these alternate use-cases, this paper compels further research and the comparison of radar point-cloud extractors used for tasks such as radar-based localization, simultaneous localization and mapping, and object detection.

\section{CONCLUSION}
This paper presents the first study examining the impact of different extractors on the overall performance of an ICP-based radar odometry pipeline. Our curated list of 13 extractors shows that optimizing extractors can significantly impact odometry performance, sometimes more than tuning the back-end algorithm. Our overall recommendation is to use the simple $K$-strongest extractor. It generalizes well across datasets and can be easily tuned for optimal performance on an arbitrary point-cloud-based odometry pipeline. This paper underscores the critical role of extractor optimization in enhancing odometry performance and compels further research to compare extractor performance in different contexts and for alternate navigation tasks.


\printbibliography

\end{document}

%% file: preamble.tex
\usepackage[l2tabu,orthodox]{nag}

\usepackage[normalem]{ulem}

\usepackage[
    backend=biber,
    style=ieee,
    sorting=none,
    natbib=true,
    doi=false,
    isbn=false,
    url=false,
    eprint=false,
    maxcitenames=1,
    mincitenames=1,
]{biblatex}

\usepackage[pdftex,colorlinks]{hyperref}

\usepackage[printonlyused]{acronym}

\usepackage{siunitx}
\usepackage[all]{nowidow}

\usepackage[dvipsnames]{xcolor}

\usepackage{tikz}
\usetikzlibrary{arrows}
\usetikzlibrary{arrows.meta}
\usepackage{makecell}

\usepackage{lipsum}

\usepackage{xspace} 

\usepackage{epstopdf}

\usepackage{import}

\usepackage{booktabs}
\usepackage{hhline}
\usepackage{tabularx}
\usepackage{multirow, multicol}

\usepackage{rotating}

\usepackage{amssymb,amsfonts,amsmath,amscd}

\usepackage{bm}

\newcommand{\bbm}{\begin{bmatrix}}
\newcommand{\ebm}{\end{bmatrix}}

\newcommand{\ignore}[1]{}

\newcommand{\bma}[1]{\left[\begin{array}{#1}}
\newcommand{\ema}{\end{array}\right]}

\DeclareMathAlphabet{\mbf}{OT1}{ptm}{b}{n}

\def\fdotb{{\raisebox{-0.6ex}{ \kern0.2ex\raisebox{0.8ex}{\tiny $\hspace*{-1ex}\circ$}}}}
\def\fddotb{{\raisebox{-0.6ex}{ \kern0.2ex\raisebox{0.8ex}{\tiny $\hspace*{-1ex}\circ\circ$}}}}

\newcommand{\utimes}{ {\raisebox{-0.6ex}{ \kern-1.0ex\raisebox{0.6ex}{ \small $\mathsf{v}$}}} } %
\newcommand{\beq}{\begin{equation}}
\newcommand{\eeq}{\end{equation}}
\newcommand{\bdis}{\begin{displaymath}}
\newcommand{\edis}{\end{displaymath}}
\newcommand{\beqarray}{\begin{eqnarray}}
\newcommand{\eeqarray}{\end{eqnarray}}
\newcommand{\beqarraynn}{\begin{eqnarray*}}
\newcommand{\eeqarraynn}{\end{eqnarray*}}

\DeclareMathAlphabet{\mbf}{OT1}{ptm}{b}{n}

\usepackage{balance}
\usepackage[T1]{fontenc}